%% file: main.tex
\documentclass[letterpaper, 10 pt, journal, twoside]{IEEEtran}

\usepackage{cite}
\usepackage{xcolor}
\usepackage{balance}
\usepackage[pdftex]{graphicx}
\usepackage{amsmath}
\usepackage{amsfonts} 
\newcommand{\subfig}[1]{\textit{\footnotesize{#1}}}

\usepackage{algorithmic}
\usepackage{cleveref} 

\newcommand{\name}{\textsc{GSON}}
\newcommand{\figref}[1]{Fig.~\ref{#1}}

\begin{document}

\title{\name{}: A Group-based Social Navigation Framework with Large Multimodal Model}

\author{
Shangyi~Luo$^{1}$,
Peng~Sun$^{1}$,
Ji~Zhu$^{1}$,
Yuhong~Deng$^{2}$,
Cunjun~Yu$^{2}$,
Anxing~Xiao$^{2}$,
Xueqian~Wang$^{1}$
\thanks{Manuscript received: April, 7, 2025; Revised June, 26, 2025; Accepted July, 21, 2025.}
\thanks{This paper was recommended for publication by Editor Markus Vincze upon evaluation of the Associate Editor and Reviewers' comments.
This work was supported by the National Key  R\&D Program of China (2022YFB4701400/4701402), and National Natural Science Foundation of China under Grant 62293545. \textit{(Shangyi Luo, Peng Sun, and Ji Zhu are co-first authors.) (Corresponding authors: Anxing Xiao; Xueqian Wang.)}} 
\thanks{$^{1}$Center for Artificial Intelligence and Robotics, Tsinghua  Shenzhen  International Graduate School, Shenzhen, China. Correspondence to: {\tt\small wang.xq@sz.tsinghua.edu.cn}
}%
\thanks{$^{2}$School of Computing, National University of Singapore, Singapore. Correspondence to: {\tt\small anxingx@comp.nus.edu.sg}}%
\thanks{Digital Object Identifier (DOI): see top of this page.}
}

\markboth{IEEE Robotics and Automation Letters. Preprint Version. Accepted July, 2025}
{Luo \MakeLowercase{\textit{et al.}}: GSON: A Group-based Social Navigation Framework with Large Multimodal Model}

\maketitle

\begin{abstract}
With the increasing presence of service robots and autonomous vehicles in human environments, navigation systems need to evolve beyond simple destination reach to incorporate social awareness. This paper introduces \name{}, a novel group-based social navigation framework that leverages Large Multimodal Models (LMMs) to enhance robots' social perception capabilities. Our approach uses visual prompting to enable zero-shot extraction of social relationships among pedestrians and integrates these results with robust pedestrian detection and tracking pipelines to overcome the inherent inference speed limitations of LMMs. The planning system incorporates a mid-level planner that sits between global path planning and local motion planning, effectively preserving both global context and reactive responsiveness while avoiding disruption of the predicted social group. We validate \name{} through extensive real-world mobile robot navigation experiments involving complex social scenarios such as queuing, conversations, and photo sessions. Comparative results show that our system significantly outperforms existing navigation approaches in minimizing social perturbations while maintaining comparable performance on traditional navigation metrics. 
\end{abstract}


\IEEEpeerreviewmaketitle

\input{sections/introduction}

\input{sections/relatedworks}

\input{sections/formulation}

\input{sections/method}

\input{sections/experiments}
\input{sections/conclusion}

{
    \balance
    \bibliographystyle{IEEEtran}
    \bibliography{IEEEabrv, bibliography}
}

\end{document}

%% file: sections/introduction.tex
\section{Introduction}

\begin{figure}[t] 
\centering
\includegraphics[width=.95\linewidth]{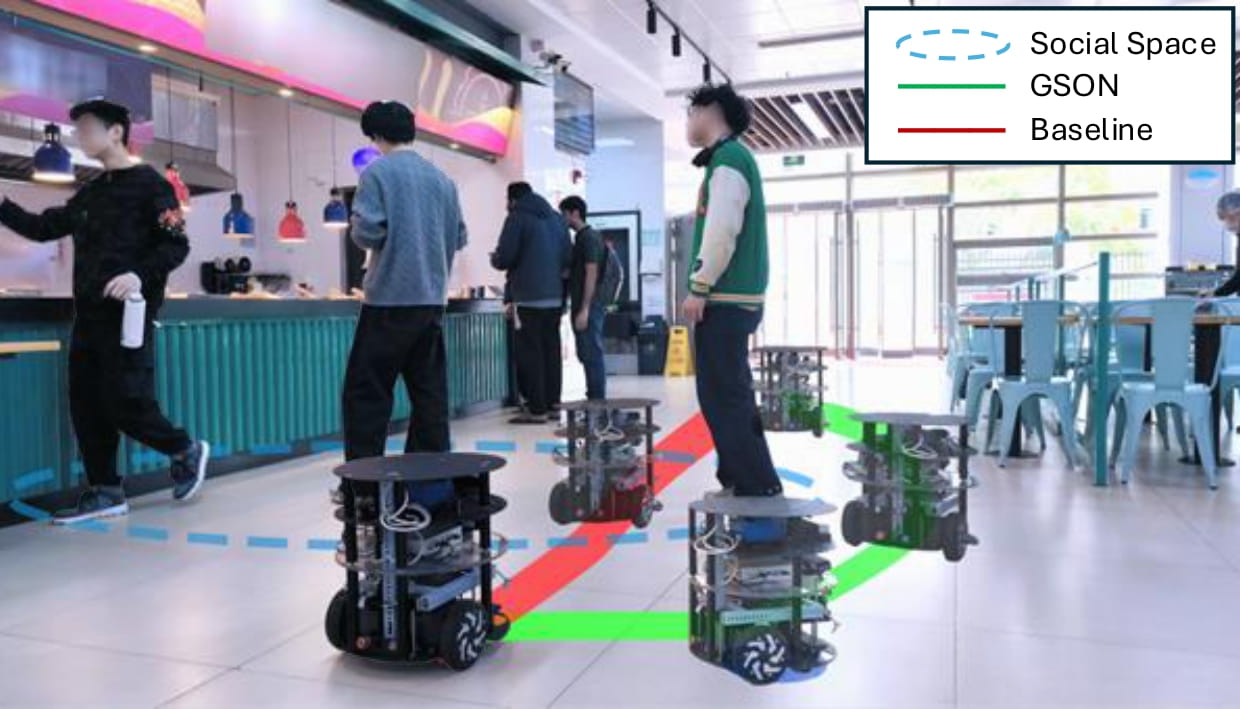}
\caption{
Example of a social navigation scenario: 
Rather than walking between two individuals in a queue, the robot correctly identifies the social context and navigates around the queue to avoid disturbance.
}
\label{fig:teaser}
\vspace{-0.7cm}
\end{figure}

\IEEEPARstart{T}{he} growth of service robots has driven significant research on autonomous systems capable of navigating human-centered environments~\cite{thrun2000probabilistic, xia2023collaborative, francisPrinciplesGuidelinesEvaluating2023}. However, a critical gap exists in current navigation systems: while they excel at trajectory prediction and obstacle avoidance~\cite{ferrer2014proactive, chen2017socially, huber2022fast, chen2023quadruped, huber2023avoidance}, they often fail to recognize and respect complex social contexts within crowds, such as photography sessions or queuing behaviors, as illustrated in~\figref{fig:teaser}. 
In the broader context of social robot navigation \cite{mavrogiannis2023core, singamaneni2024survey}, the goal is not only for the robot to reach its destination but also to interact appropriately with humans without degrading their experience. In this work, we focus on \textit{developing a robot navigation system that can understand and respect contextual social interactions}, which we represent as social groups. These social groups represent meaningful patterns of semantic interaction among individuals within the crowd.

The core issue in building such a socially aware navigation system is how to accurately \textit{identify} contextual social interactions in dynamic and unpredictable crowds, and \textit{exploit} these social interactions to guide the planning system.
For perception, accurately identifying contextual social interactions in dynamic, unpredictable crowds requires semantic understanding beyond simple proximity or velocity patterns. Traditional methods based on predefined rules or in-domain learning~\cite{ferrer2014proactive, chen2017socially, gupta2018social, wang2020group, taylor2020robot, wangGroupbasedMotionPrediction2022, jahangard2023real, lu2025group} struggle in open-world environments due to data scarcity and high visual variability. Regarding planning, generating socially appropriate motions that respect these interactions demands planning at the right abstraction level—traditional local planners often fall into local minima due to short planning horizons~\cite{fox1997dynamic, van2011reciprocal, panagou2014motion, rosmann2015timed}, while global path planners cannot adapt quickly to dynamic social interactions.
In contrast to modular systems, end-to-end solutions aim to bypass the traditional perception-planning decomposition \cite{ tai2018socially, xie2021towards,hirose2023sacson, nguyen2023toward}. However, these approaches often overfit demonstrations and struggle to interpret complex contextual information. More recently, some works have explored the use of large multimodal models (LMMs) to select sampled trajectories \cite{narasimhan2024olivia} or high-level directions\cite{song2024vlm, payandeh2024social, kong2025autospatial} to improve social awareness without explicit social interactions modeling. However, these methods are inefficient as they require querying the LMM at every planning step, and the resulting social behaviors are often not interpretable. 

\begin{figure*}[!t] 
\centering
\includegraphics[width=0.9\linewidth]{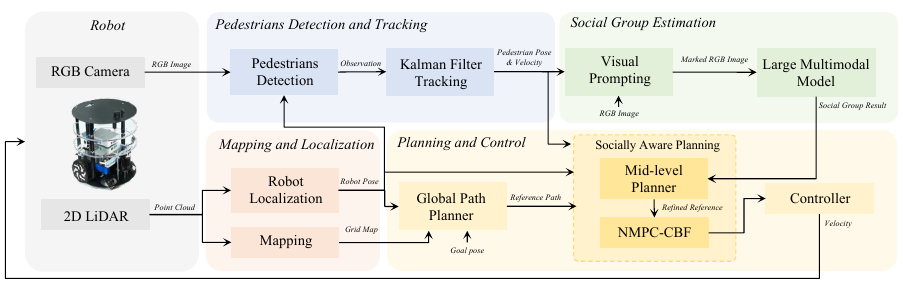}
\caption{ Overall Framework for GSON:
GSON integrates RGB cameras and 2D LiDAR data for dynamic crowd perception. The large multimodal model allows group estimation and enables the planning system to perform socially-aware planning for navigation.
}
\label{fig:framework}
\vspace{-0.7cm}
\end{figure*}

To systematically address these challenges, we propose \name{} that combines the strengths of Large Multimodal Models (LMMs) with a robust navigation system. Our key insight is to utilize the visual understanding ability of the Large Multimodal Model (LMM) to enable zero-shot reasoning of social groups. 
The LMM serves as an auxiliary to our perception pipeline, which supports reliable detection and tracking, to enhance the estimation of social groups. Our design utilizes LMMs only for low-frequency semantic understanding, while high-frequency tracking and planning are handled without querying large models, leading to a more practical sense.
In addition, we adapted a mid-level planner as a bridge between global path planning and local motion planning, enabling a balance between global awareness of social interactions and reactive responsiveness.

Overall, we develop a group-based social navigation framework \name{} that can identify and exploit contextual social interactions. We conduct extensive experiments with several daily social interaction scenarios in the real world. Comparative experiments indicate that \name{} outperforms all baseline methods in terms of minimizing perturbations to the social group.

%% file: sections/relatedworks.tex
\section{Related Work}
\label{sec:related}
\subsection{Social Group Detection in the Crowd.}
Significant research has focused on social group detection in crowded pedestrian environments, which can be approached from both exo-centric and ego-centric perspectives.
The exo-centric approach \cite{trautmanRobotNavigationDense2013} employs external sensors to provide a comprehensive top-down view for crowd analysis. Various methods have been proposed for exo-centric group detection to support motion tracking \cite{zhu2014crowd, wang2018detecting}, trajectory prediction\cite{wang2020group,fangneuralized}, group activity recognition\cite{chappa2023spartan, kim2024towards}, and crowd analysis \cite{solera2015socially, li2017multiview, li2022self}. 
However, this approach has two key limitations: it relies on external sensors and offline processing, and it often overlooks the rich semantics of crowd interactions by focusing mainly on motion patterns or domain-specific overfitting.
For ego-centric group detection, prior approaches to crowd grouping have employed clustering based on motion patterns \cite{luber2013multi, taylor2020robot, wangGroupbasedMotionPrediction2022} and distance \cite{ wang2022group,taylor2022regroup, wangGroupbasedMotionPrediction2022}, or
learning group information in a specific dataset using graph neural network \cite{jahangard2023real, lu2025group}, predicting cost function\cite{eirale2024learning}, or leveraging F-formation theories\cite{barua2024enabling}.
While these methods excel at robot-centric perception, they still face limitations in capturing semantic context and tend to overfit specific scenarios.
In \name{}, we leverage the zero-shot visual reasoning capabilities of the Large Multimodal Model (LMM) to enable semantic reasoning of social groups given ego-centric observation. 

\subsection{Social Robot Navigation}
Previous studies have proposed various methods to improve collision avoidance in robot navigation \cite{van2011reciprocal, chen2017socially, huber2022fast, huber2023avoidance}. With the increasing number of mobile robots in human environments, there's a need beyond collision avoidance to improve social awareness.
Traditionally, researchers have applied physics-inspired approaches such as the Social Force Model \cite{helbing1995social} to robot navigation \cite{ferrer2014proactive, biswas2022socnavbench}. However, SFM-based methods only implicitly represent social awareness as pedestrian motion and suffer from limited generalization due to parameter selection.
Group-based navigation strategies integrate dynamic human grouping into social navigation. Methods that combine group information with model-based planners \cite{luber2013multi, taylor2022regroup, wangGroupbasedMotionPrediction2022} or reinforcement learning controllers \cite{katyal2022learning, lu2025group} show that representing human groups improves robot safety and social awareness.
However, most group-based methods define groups only in terms of position and velocity, ignoring the semantic social interaction between pedestrians. 
Another line of research uses imitation learning \cite{tai2018socially, xie2021towards,hirose2023sacson, nguyen2023toward} to implicitly learn social awareness from demonstrations. However, due to the data scarcity problem, the learned policies capture only limited behaviors in the dataset.
Recent advances in foundation models have integrated large multimodal models to generate high-level actions \cite{payandeh2024social, kong2025autospatial}, score trajectories \cite{narasimhan2024olivia} or directions \cite{song2024vlm}, and distill cost into imitation learning \cite{elnoor2025vi}. These methods are inefficient, requiring LMM queries at every planning step, with often uninterpretable social behaviors.
Our approach uses LMMs only for low-frequency semantic social group detection when new pedestrians appear in observation, handling high-frequency tracking and planning without querying large models, creating a more practical system.

%% file: sections/formulation.tex
\section{Problem Formulation}
We consider the problem of social robot navigation in a 2D Euclidean space \( \mathbb{R}^2 \). 
This space is populated by \( n \) dynamic pedestrians $H_i$ with a specific social structure. We consider the social structure as \( m \) social groups $G_j$. 
The robot is initialized with a 2D occupancy map \(M \) at the start position $S_s$ and a target position $S_g$ in the map. Given the 2D lidar point cloud $P^t$ and the RGB image observation $I^t$, the goal is to enable the robot to safely navigate to the goal position while minimizing the time it spends trespassing on any social group. The robot must infer the social groups of pedestrians from the limited observation and generate an appropriate trajectory $T$.
We focus on designing the method to estimate the social group with the aid of the Large Multimodal Model and integrate the estimated result into the navigation system.

%% file: sections/method.tex
\section{Method}
\subsection{Overview}
The overall architecture of \name{} is illustrated in~\figref{fig:framework}.
The robot is equipped with two RGB cameras, a 2D lidar sensor, and a mobile base with a differential drive. The system has two critical modules: the social group estimation module and the socially aware planning module. The social group estimation module takes as input the current observation RGB image $I^t$ (formed by merging images from multiple cameras), 2D point clouds $P^t$, and the current robot pose $S^t$ from the SLAM module. It generates the estimated social group score $\hat{G}$, quantifying the likelihood that individuals belong to the same social group. 
The estimated social group score is fed into the socially aware planning modules, which consist of a global path planner, a mid-level planner, and a local motion planner. The global path planner computes a collision-free path from the robot's starting position to the target position, taking into account the environment information and static obstacles such as walls, furniture, and other permanent structures. The mid-level planner utilizes the estimated social group information and generates intermediate references that guide the local planner in navigating through the crowd while minimizing the risk of intruding upon the social groups. The local motion planner then ensures precise and safe navigation based on the intermediate references provided by the mid-level planner.

\begin{figure}[t]
    \centering
    \includegraphics[width=0.9\linewidth]{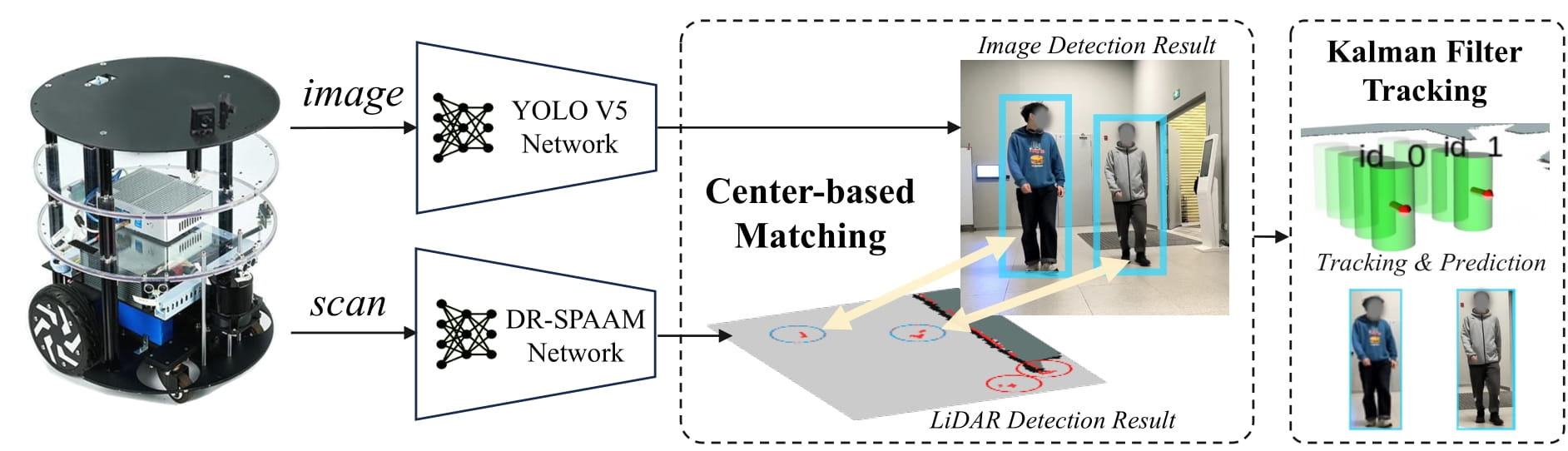} 
    \caption{The framework of pedestrians detection and tracking: It combines 2D LiDAR-based foot detection with human detection from RGB images}
    \label{fig:detect}
    \vspace{-0.6cm}
\end {figure}

\subsection{Social Group Estimation with Large Multimodal Models}
The core idea of \name{} is to model contextual social interactions in dynamic and unpredictable crowds using social groups, which then guide the planning system. In this context, a social group refers to two or more individuals who are interacting, conversing, or moving together with shared intentions or social connections. To identify these social groups, we first build a robust pipeline for pedestrian detection and tracking. After obtaining detection and tracking results of pedestrians, identifying social groups requires commonsense knowledge about social interaction. Therefore, we leverage the multi-modal reasoning capabilities of the LMM and employ visual prompting to identify social groups.

\noindent\textbf{Pedestrian Detection and Tracking:} We design a robust pipeline for pedestrian detection and tracking. Given the point cloud data $P^t$ of the environment, we first use an off-the-shelf model proposed in~\cite{jia2020dr} for preliminary pedestrian detection. This preliminary detection provides potential pedestrian locations in the 3D point cloud but may contain false positives.

To improve accuracy and remove noise, we combine information from RGB images using YOLO-V5~\cite{yolov5}. Specifically, the preliminary detected pedestrian coordinates $p_i^{3D} = [x_i^{\prime}, y_i^{\prime}, z_i^{\prime}]$ from the point cloud are projected into the camera coordinate system. We then compare these projections with the bounding box results from YOLO-V5, retaining only those detections that align with YOLO-V5's pedestrian identifications. 
For each YOLO-V5 detection box $B_j$ (where $j \in \{1, \dots, N\}$), we define the final selected coordinate $h_j^*$ by:
\begin{equation}
    h_j^* = \arg\min_{h_i \in \mathcal{H}_j^{proj}} \left\| \pi(h_i) - \mathbf{c}_j \right\|_2,
\end{equation}
where $\mathcal{H}_j^{proj} = \big\{ p_i^{3D} \, \big| \, \pi(p_i^{3D}) \in B_j \big\}$ is the set of point cloud coordinates whose projections fall within the bounding box $B_j$, $\pi: \mathbb{R}^3 \to \mathbb{R}^2$ denotes the projection from 3D point cloud coordinates to 2D image coordinates,$\mathbf{c}_j = (c_j^x, c_j^y)$ is the center coordinates of bounding box $B_j$, $\| \cdot \|_2$ represents the Euclidean distance.

After this filtering process, we obtain refined pedestrian positions in 2D coordinates $\mathbf{p}_i = [x_i, y_i]$, where the $z$ dimension is omitted as we focus on the ground plane for navigation purposes. Finally, we apply the Hungarian algorithm for data association and Kalman filtering for estimating and predicting the position of each pedestrian with a unique tracking ID, enabling robust tracking for subsequent social group detection. \textcolor{black}{Each pedestrian is modeled using a constant velocity motion model, represented by a state vector $\mathbf{s}_k = [x_k, y_k, \dot{x}_k, \dot{y}_k]^\top$, which encodes both position and velocity in 2D space.}
At each timestep, Kalman filters first predict target positions \textcolor{black}{using linear constant-velocity dynamics}, then the Hungarian algorithm optimally matches these predictions with new detections by minimizing the total Euclidean distance between corresponding points. Unmatched detections initialize new tracks with zero initial velocity, while tracks failing to match consecutively for several frames are terminated to prevent ghost tracking.

This pipeline enables robust pedestrian detection and tracking, providing bounding boxes and tracking IDs necessary for subsequent social group detection. \textcolor{black}{In addition to supporting group detection, the tracking results—specifically each pedestrian’s estimated position and velocity—serve as crucial input to the planning module. These motion states help the planner reason about group formations and predict pedestrian movement, enabling proactive, socially aware trajectory generation that avoids disrupting group interactions.}

\begin{figure}[!t]
    \centering
    \includegraphics[width=0.9\linewidth]{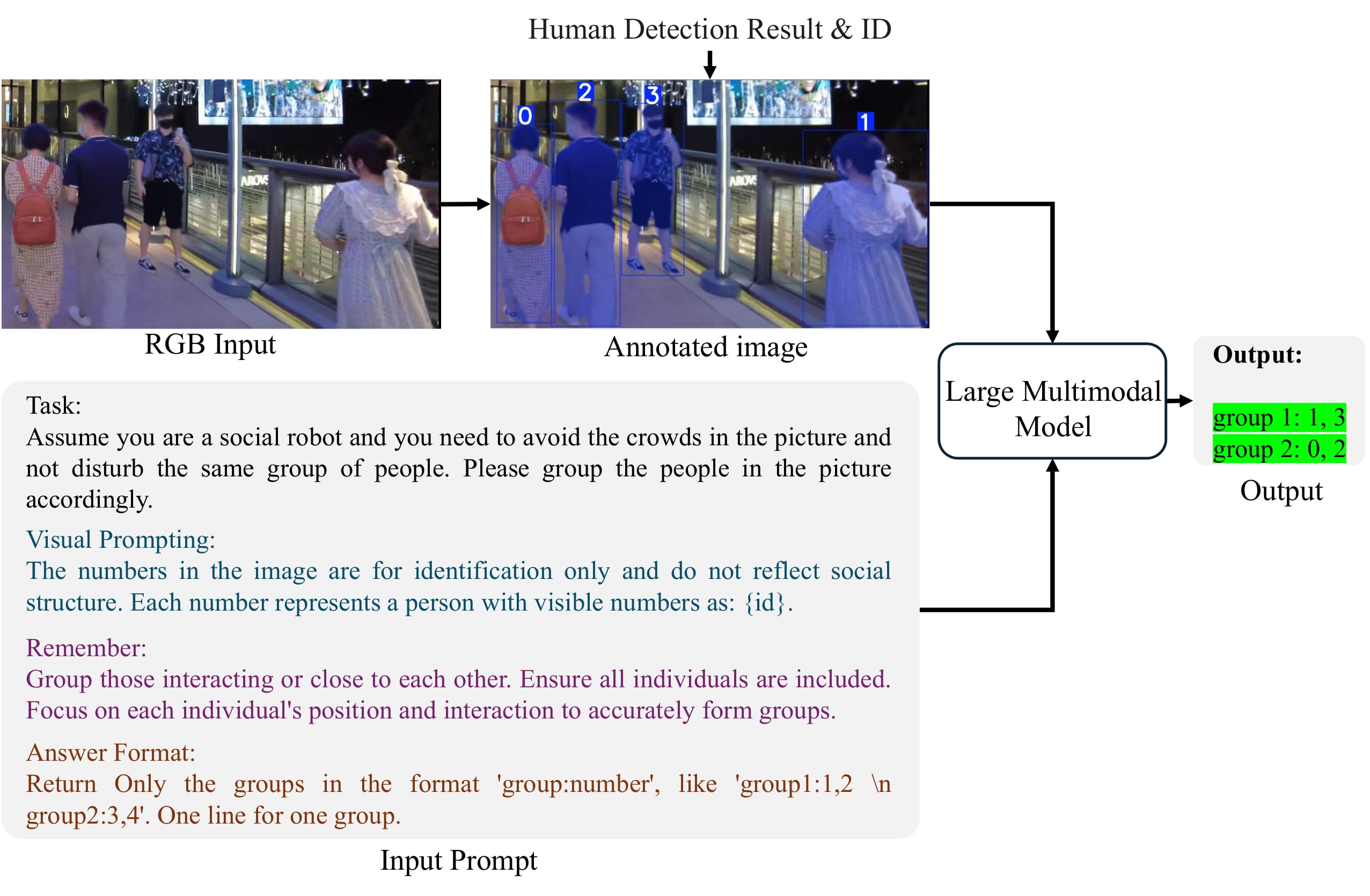}       
    \caption{With visual prompting, the LMM identifies social interactions among people and infers social groupings for downstream planning modules.}
    \label{fig:LMM}
    \vspace{-0.7cm}
\end {figure}

\noindent\textbf{Social Group Detection with Visual Prompting:}  We leverage a LMM~\footnote{specifically gpt-4o-2024-08-06} to identify social groups within crowds by analyzing human behaviors and interactions. Our approach consists of three main steps: image annotation, LMM querying, and social group relationship extraction as shown in Fig.~\ref{fig:LMM}. First, given bounding boxes and tracking IDs of pedestrians, 
We use the mask that results from YOLO-V5, creating an annotated image where each individual is marked with their bounding box, segmentation mask, and unique tracking ID $h_i$. 
These visual marks enhance the LMM's reasoning. Second, we query the LMM with a structured input consisting of the annotated image and a text prompt. In the text prompt, we provide the \textcolor{black}{LMM} with contextual information, including the expected behavior of the LMM, the format of queries and responses, and cues to assist the \textcolor{black}{LMM} in accurately identifying social groups. Finally, we process the LMM's output using a parsing script that extracts the predicted social group relationships. These relationships are then integrated into our tracking system's data structure, maintaining an up-to-date record of social groups over time.
Formally, we can express this process as:
\begin{equation}
    G = \textrm{LMM}(A(I,\mathcal{H}^{track}),T),
\end{equation}
where $I$ denotes the raw input image frame, $\mathcal{H}^{track} = \{p_i\}_{i=1}^N$ represents the filtered tracking states, and $A$ is our annotation function that applies masks, bounding boxes, and IDs to the image, $T$ is the text prompt, and $G$ is the resulting social group assignments. Our approach enables the system to identify social groups by analyzing visual cues and spatial relationships between individuals.

Due to the computational constraints of LMM, we implement a keyframe-based approach to reduce the frequency of LMM queries while maintaining accurate social group tracking. Instead of analyzing every frame, we maintain a keyframe that captures the most informative scene within a 1-second window - specifically, the frame containing the highest number of visible people. Each keyframe stores both the image and corresponding metadata (position and velocity) for all detected individuals. The LMM is only queried when this keyframe is updated, which occurs either when a more representative frame is found or when the current keyframe expires after 1 second.

\subsection{Socially Aware Planning Module}
To leverage social group identification results for guiding robot planning, we propose a socially aware planning module, which consists of three levels: global path planning, mid-level planning, and local motion planning. \par

\noindent\textbf{Global Path Planning:} The first step of our planning module is to compute a collision-free path from the robot's current position to the target location. We begin by inflating the occupancy map $M$ to account for the robot's size and safety buffer, resulting in the grid-based cost map $C$. This inflated cost map provides a conservative estimate of obstacle boundaries. We then implement the A* algorithm on $C$, further optimized with Jump Point Search (JPS) \cite{harabor2011jps}. JPS improves the efficiency of path planning in large-scale grid environments by reducing unnecessary node explorations. The generated global path $R$ serves as the reference for the mid-level and local planners.

\noindent\textbf{Mid-level planning:}
Given the global reference path $R$, we introduce a mid-level planning algorithm that leverages social groups to provide socially aware guidance in the planning module.

\begin{figure}
    \centering
    \includegraphics[width=0.9\linewidth]{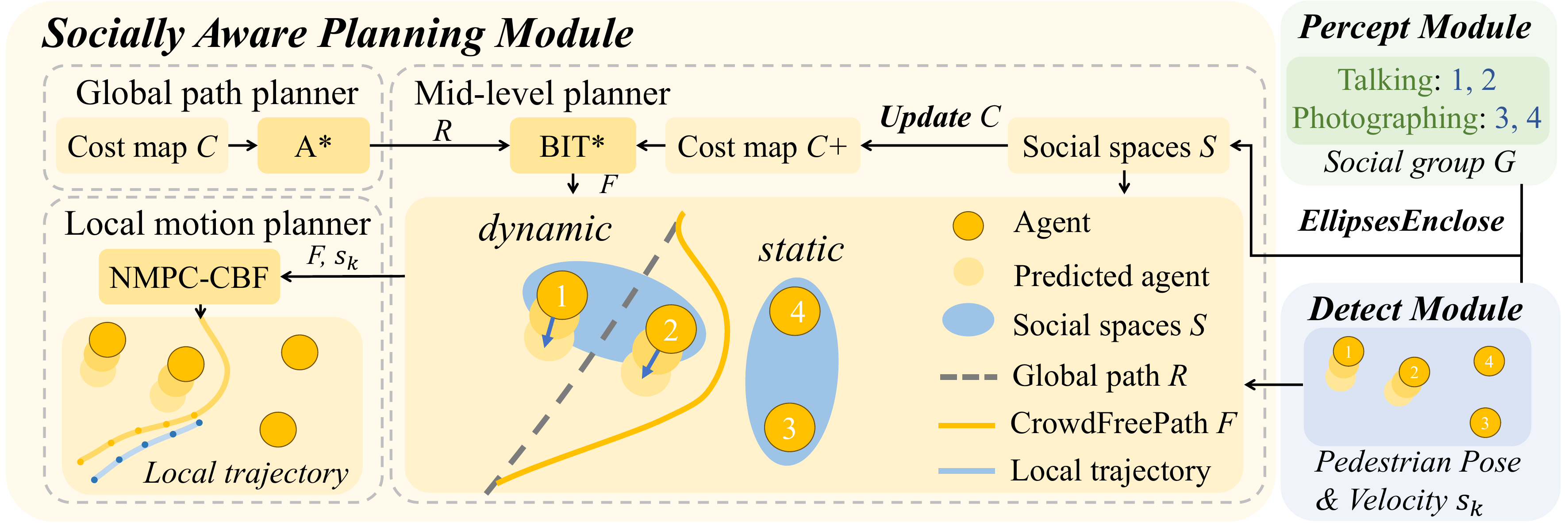}       
    \caption{\textcolor{black}{The framework of socially aware planning modules.}}
    \label{fig:planning}
    \vspace{-0.7cm} 
\end{figure}

Once the LMM module has produced the social group assignments $G$, we proceed to estimate the spatial extent of each group. Specifically, we compute the convex hull of each group, denoted as $G'$, to represent the minimal boundary enclosing all its members. To further approximate the shape and spatial influence of each group, we fit an ellipse to the boundary defined by $G'$. The resulting enclosed two-dimensional regions are defined as social spaces $S$, as shown in Fig.~\ref{fig:planning}.
\textcolor{black}{Once the global reference path $R= \{\mathbf{p}_0,\dots,\mathbf{p}_N\}$ intersects a social space $\mathcal{S}\subset C^{+}$, the robot’s current state is set as a new start $S'_s$. Let $\mathbf{p}_k$ be the last waypoint within $\mathcal{S}$. From $\mathbf{p}_k$, we move a fixed arc-length $h$ along $R$ to define a sub-goal $S'_g$. If the remaining path is shorter than $h$, set $S'_g = \mathbf{p}_N$; otherwise, find the smallest $m$ such that the arc length from $\mathbf{p}_k$ to $\mathbf{p}_m$ is at least $h$, and interpolate between $\mathbf{p}_{m-1}$ and $\mathbf{p}_m$ to locate $S'_g$. This ensures $S'_g$ lies on $R$ but outside $\mathcal{S}$, so only the segment $\overline{S'_sS'_g}$ requires re-planning.}

To compute a socially compliant trajectory under the updated cost constraints, we employ the Batch Informed Trees (BIT*) algorithm \cite{gammell2020batch}. This process yields a revised path, referred to as the \emph{CrowdFreePath} $F$, which effectively avoids traversing through socially sensitive regions and minimizes potential interactions with social groups.
The socially-aware mid-level planning incorporates social group information, enabling dynamic updates to the cost map. This ensures the robot navigates safely while avoiding significant impacts on social groups.

\noindent\textbf{Local Motion Planning:}
\textcolor{black}{Given the socially aware path $F$ from the mid-level planner and current pedestrian states (positions and velocities), the local planner computes dynamically feasible trajectories that track $F$ and ensure real-time, collision-free navigation by extrapolating pedestrian motion using a constant-velocity model.
} 
To this end, we develop a local planning framework based on Nonlinear Model Predictive Control with Control Barrier Functions (NMPC-CBF) \cite{zeng2021enhancing}, which explicitly accounts for both the positions and velocities of nearby pedestrians when computing control actions.
We define a Control Barrier Function (CBF) as follows:
\begin{equation} h^i(x) = (x - x^i_p)^2 + (y - y^i_p)^2 - d_{\text{safe}}^2, \end{equation}
where $ \mathbf{x}_p = [x_p, y_p]^T $ denotes the predicted pedestrian position based on their current velocity, and $d_{\text{safe}}$ represents a safety radius. 
The local planning problem is then formulated as a constrained NMPC optimization similar to \cite{xiao2022robotic}:
\begin{align}
\vspace{-0.3cm}
\small
    \min_{\{ \mathbf{x}_k, \mathbf{u}_k \}} & \quad \|\mathbf{x}_N - \mathbf{x}_{\text{goal}}\|_{P_f}^2 + \sum_{k=0}^{N-1} \|\mathbf{u}_k\|_{Q_u}^2  \\
    \text{s.t.} \quad & \mathbf{x}_{k+1} = f(\mathbf{x}_k, \mathbf{u}_k)  \\
    & \mathbf{x}_0 = \mathbf{x}_\text{init}  \\
    & \mathbf{x}_k \in \mathcal{X}, \mathbf{u}_k \in \mathcal{U}  \\
    & \Delta h_{\text{ob}}^i(\mathbf{x}_k, \mathbf{u}_k) + \lambda_k h_{\text{ob}}^i(\mathbf{x}_k) \geq 0 .
\end{align}
where the last constraint ensures forward invariance of the safe set defined by each pedestrian, which evolves dynamically with their predicted trajectories.

%% file: sections/experiments.tex
\section{Experiments}
In this section, we aim to investigate the following research questions through a series of experiments: \textit{(1) Can Large Multimodal Models effectively identify social groups among pedestrians? (2) How well does \name{} perform in real-world scenarios that require social awareness?
}
\begin{figure}[t]
\centering
\includegraphics[width=0.9\linewidth]{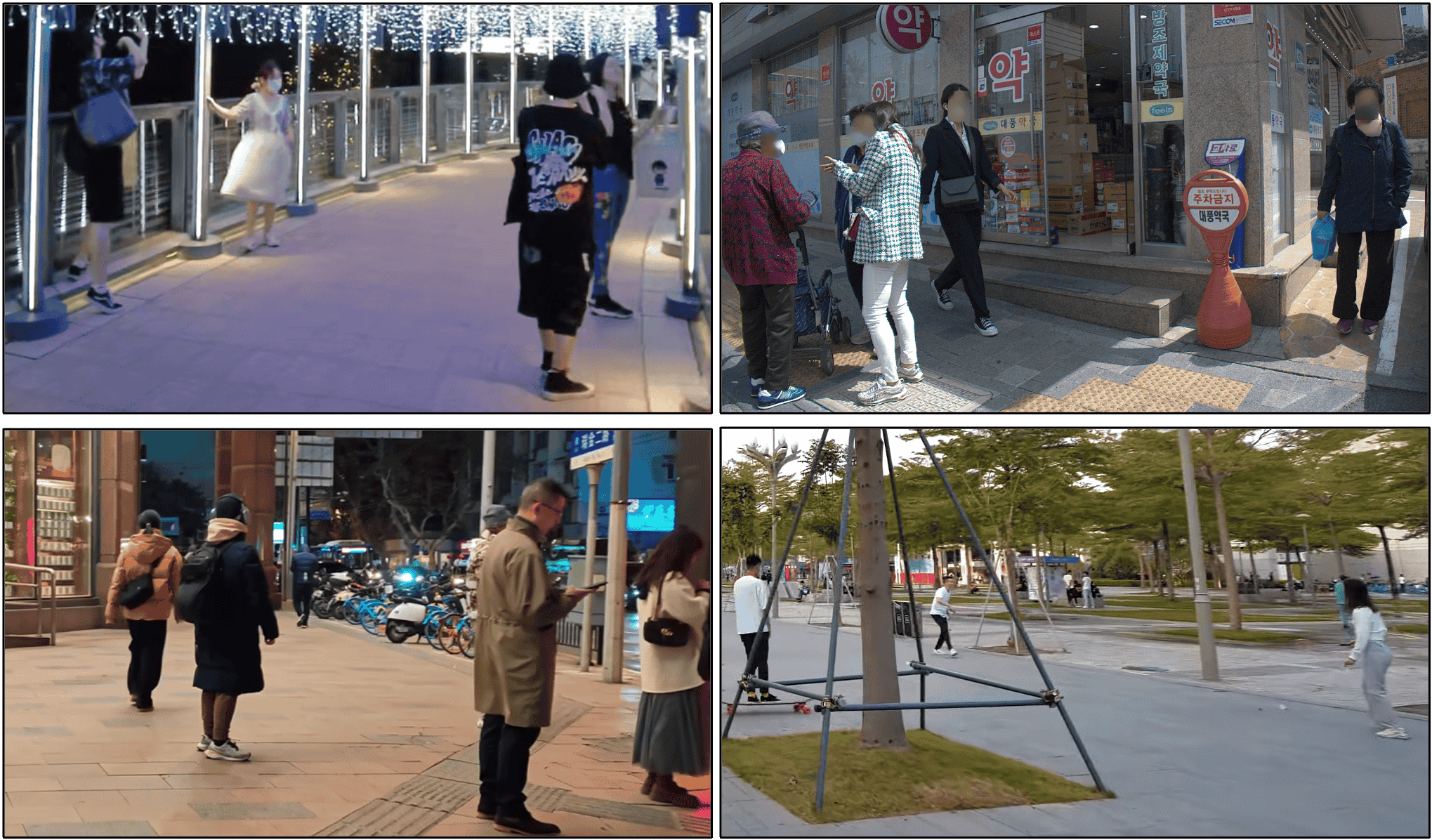}
\caption{Sampled scenario visualizations from the dataset.}
\label{figure:dataset}
\vspace{-0.6cm}
\end{figure}

\subsection{Evaluation of LMM Social Group Prediction}
To evaluate LMMs in predicting social groups from RGB images, we selected 100 scenarios from datasets~\cite{karnan2022socially, bae2023sit} and city walk videos, covering interactions such as queuing, walking, talking, and photographing. Human annotators provided group labels based on marked images, with the most agreed-upon annotations as ground truth. Example scenarios are shown in~\figref{figure:dataset}. On average, each scenario contains 6.0 pedestrians and 3.5 distinct social groups.
We evaluated six LMMs for this task: four commercial models: Gemini 2.0 Flash, Gemini 2.5 Pro, Claude 3.7 Sonnet, GPT-4o, and two open-source models: Qwen2.5-VL (32B and 7B variants). Each scenario was tested 10 times with each model. The results were classified into four categories based on the predicted groupings:
(1) \emph{Accurate}: Group matches the ground truth (preferred),
(2) \emph{Miss}: Omits some individuals from the ground-truth group (acceptable),
(3) \emph{Extra}: Includes individuals not in the ground-truth group (not preferred),
(4) \emph{Error}: Both misses and includes incorrect individuals (not acceptable).

\begin{figure}[t]
 \centering
\includegraphics[width=0.9\linewidth]{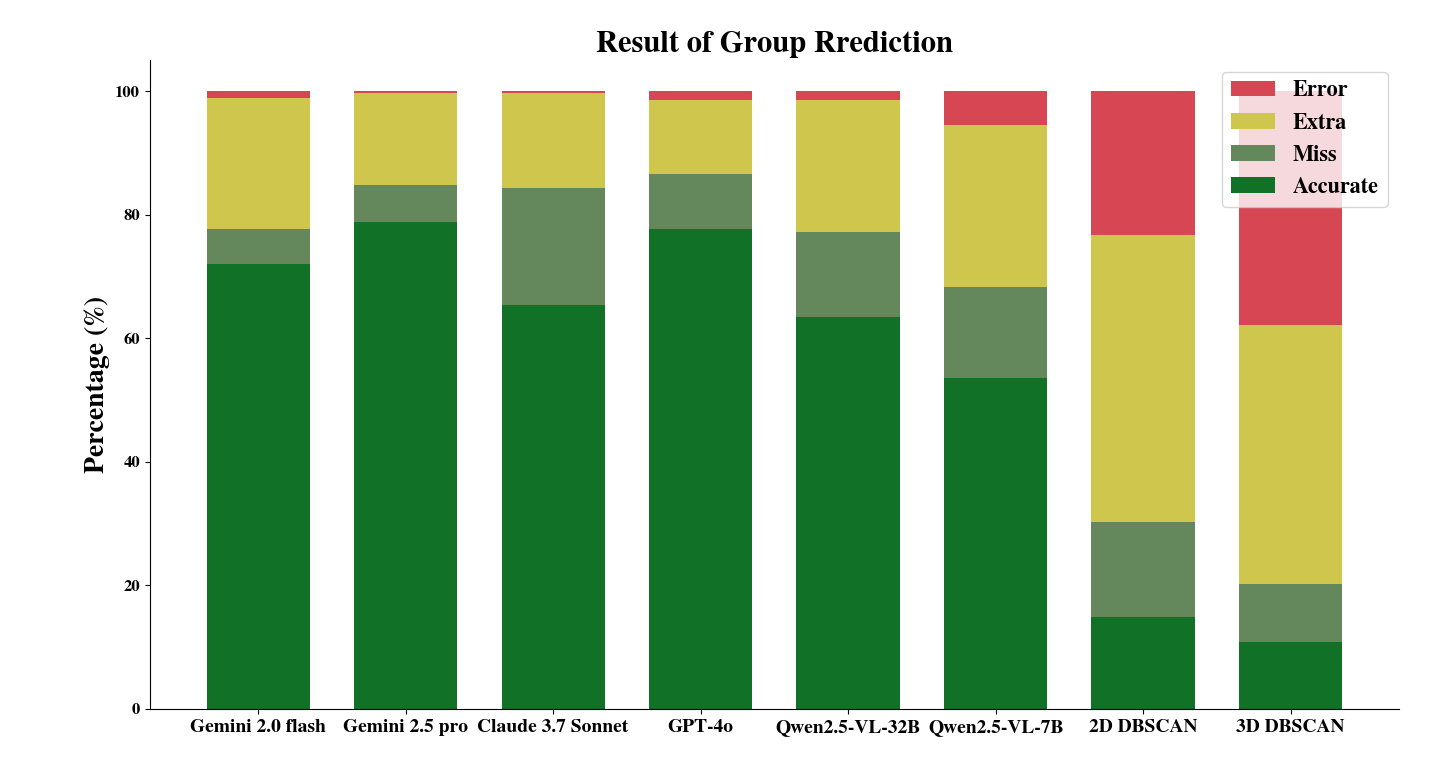}
\caption{\textcolor{black}{Grouping results of LMMs and traditional clustering in the dataset}}  \label{figure:LMMs_test}
  \vspace{-0.6cm}
\end{figure}

Fig. \ref{figure:LMMs_test} shows the results for each model. All four LMMs achieved acceptable grouping rates above 67\%, with the lowest from Qwen2.5-VL-7B (53.6\% \emph{Accurate} + 14.8\% \emph{Miss}) and the highest from GPT-4o at 86.6\% (77.8\% \emph{Accurate} + 8.8\% \emph{Miss}). Given the complexity and diversity of the dataset, these results highlight the capability of Large Multimodal Models to identify social groups among pedestrians. Inference time varies across models: Gemini 2.5 Pro and Claude 3.7 Sonnet take around 8s, GPT-4o and Gemini 2.0 Flash about 4s, Qwen2.5-VL-32B around 2s, and Qwen2.5-VL-7B is the fastest, around 1s. In our pipeline, group detection doesn't require high frequency, making GPT-4o and Qwen2.5-VL-32B well-suited for our system. 
\textcolor{black}{For comparison, we evaluated two classical clustering baselines using DBSCAN on 2D and 3D coordinates. Both yielded significantly lower accuracy (10.8–14.8\%) and higher extra grouping errors (41.9–46.5\%), underscoring the limitations of proximity-based methods and the benefits of semantic reasoning in LMMs.}

\subsection{Real-world Evaluation of \name{}}
\subsubsection{Experimental Setup}
\begin{figure}[!h]
  \centering
  \vspace{-0.2cm}
  \begin{tabular}{@{}cc@{}}
    \includegraphics[width = 0.22\textwidth]{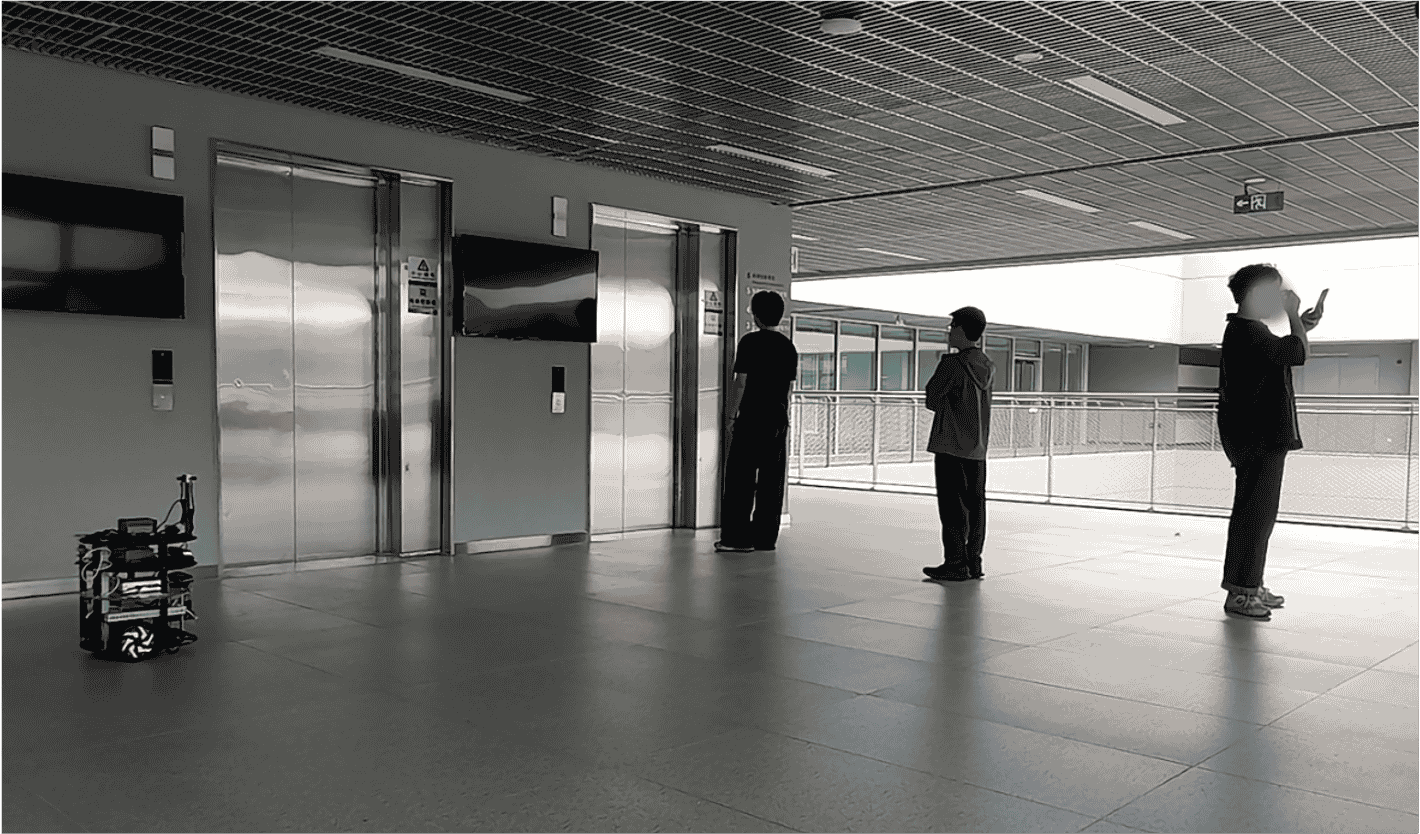}
    &
    \includegraphics[width = 0.22\textwidth]{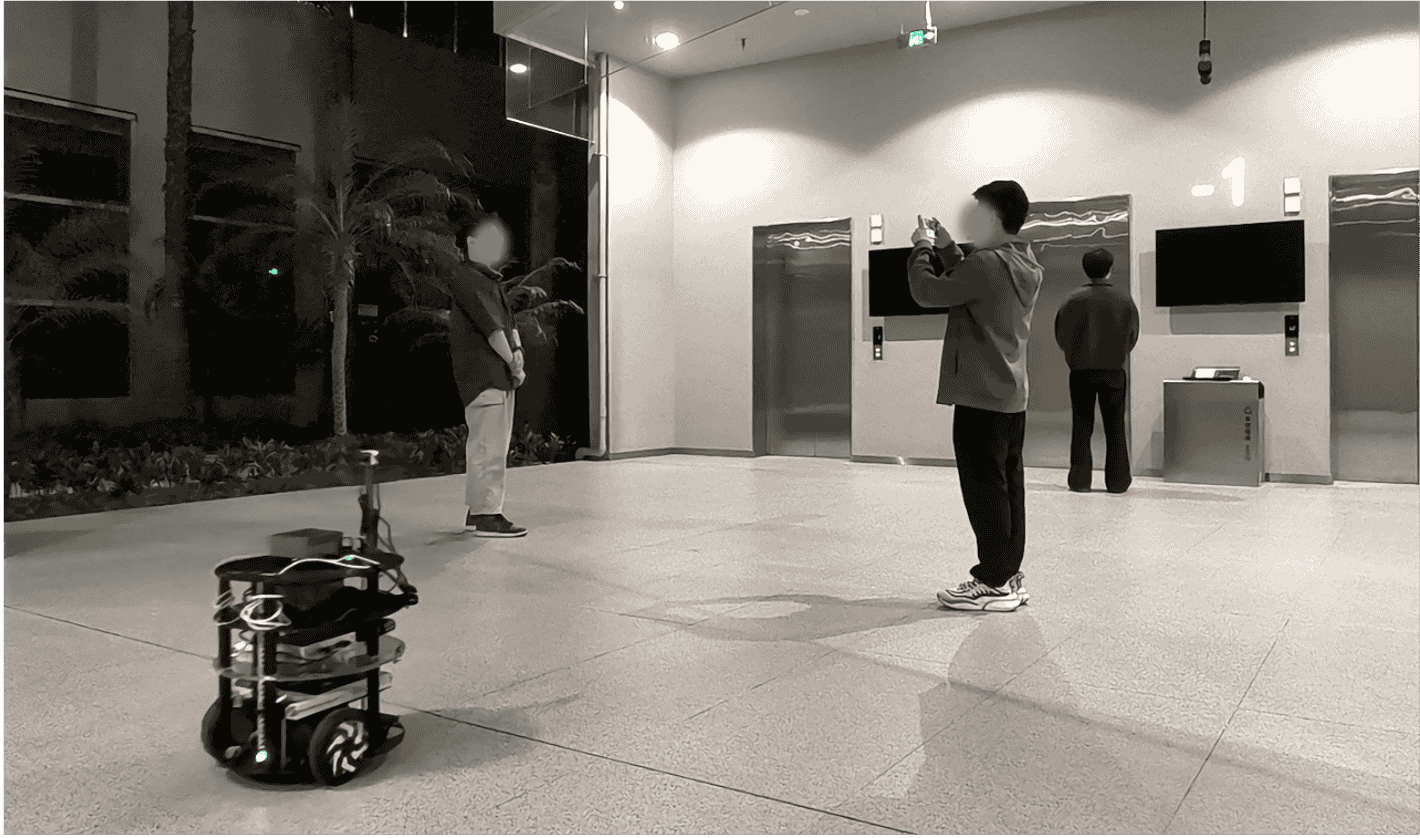} 

    \vspace{-0.1cm}
    \\
    \subfig{a) Queuing}
    &
    \subfig{b) Photographing}
    \\
    \includegraphics[width = 0.22\textwidth]{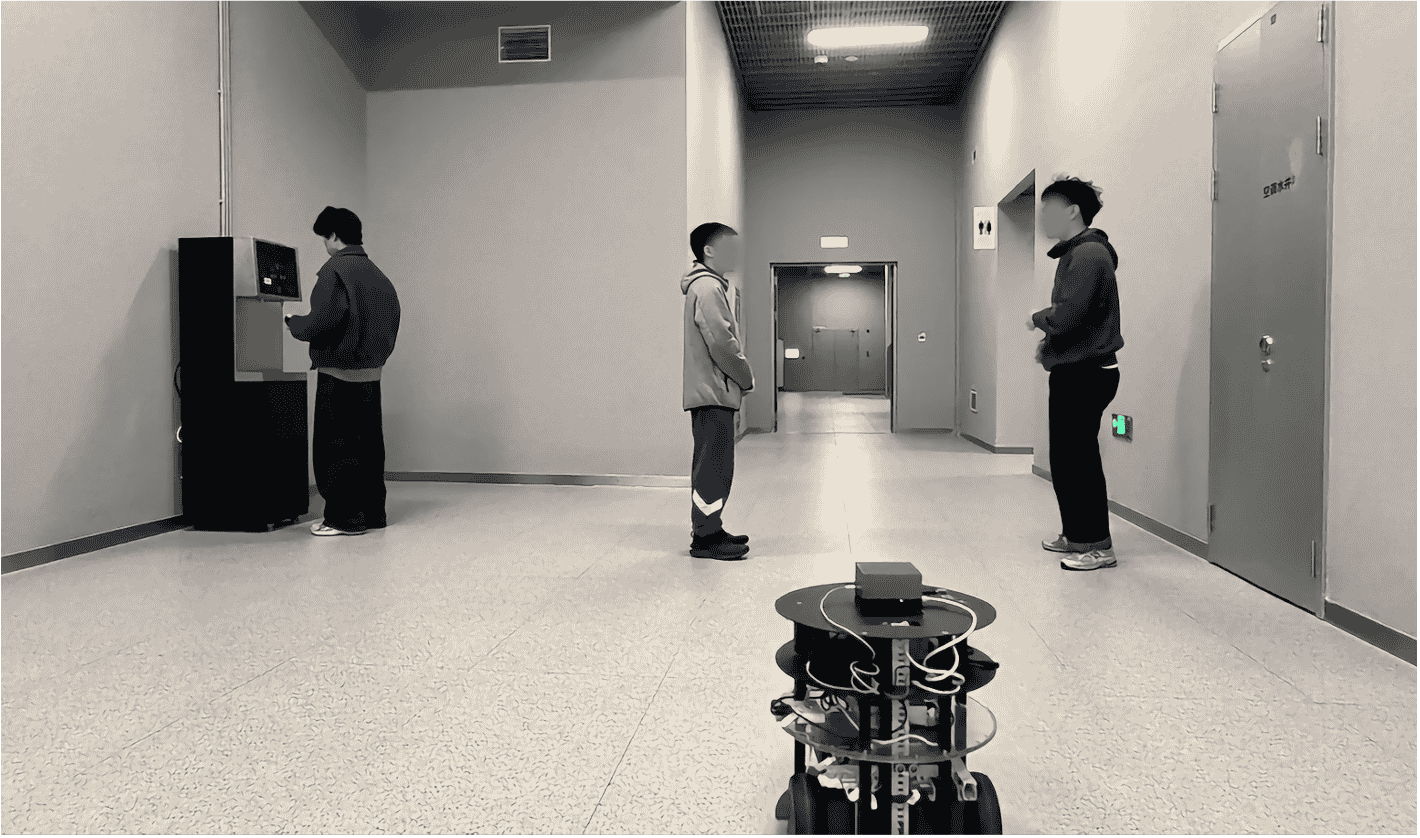}
    &
    \includegraphics[width = 0.22\textwidth]{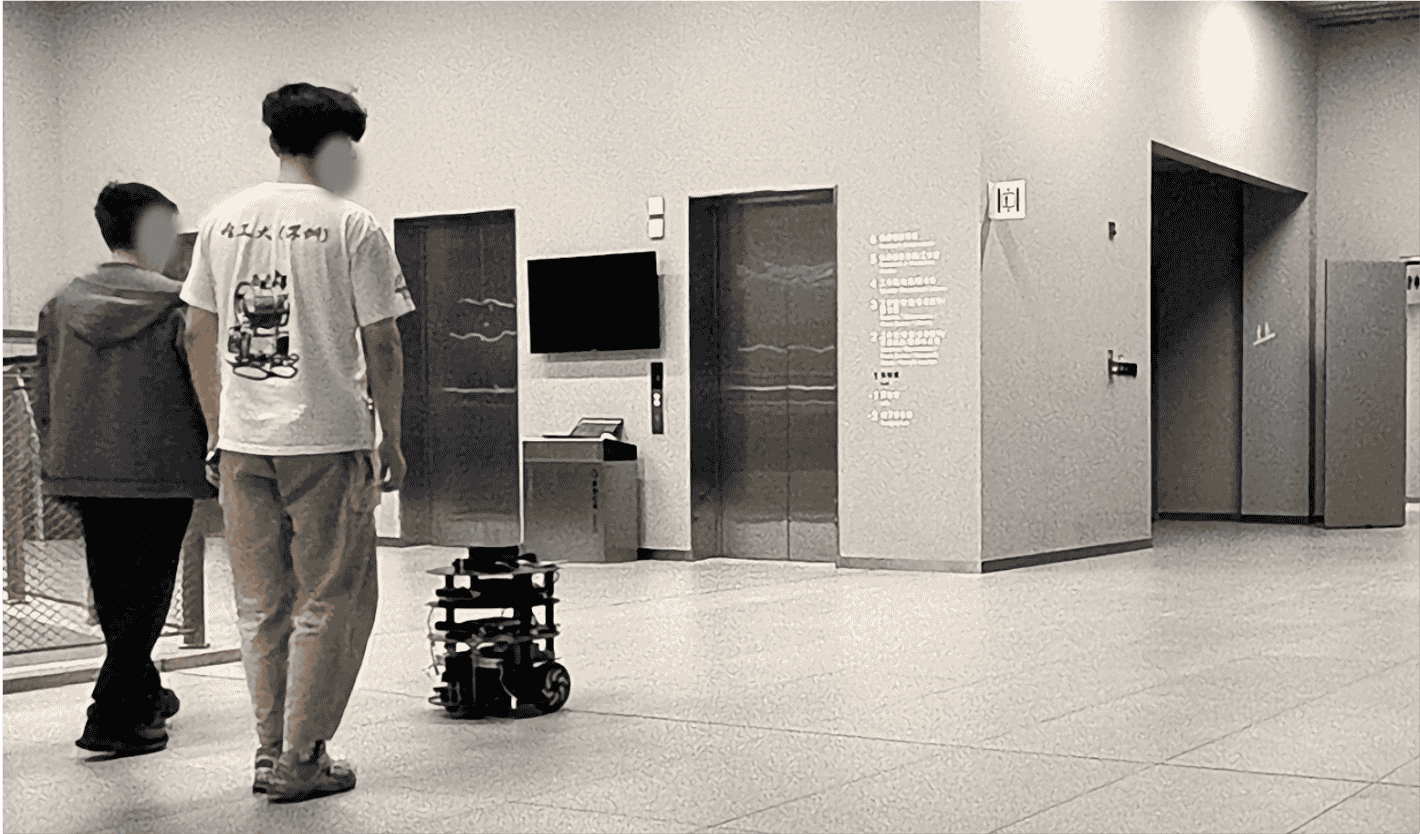}
    \vspace{-0.1cm}
    \\    
    \subfig{c) Talking}
        & 
    \subfig{d) Talking \& Walking
}
    \\ 
    \vspace{-0.4cm}
    \\
    
  \end{tabular}
  \caption{Test scenarios in real-world experiments.
}
\vspace{-0.6cm}
\label{fig:experiment-setup}
\end{figure}

We tested \name{} in real-world settings using a two-wheeled nonholonomic mobile robot equipped with a 2D LiDAR and two RGB cameras. The experiments focused on short-range navigation around people engaged in social interactions like queuing, photographing, talking, and walking, as shown in~\figref{fig:experiment-setup}. Each scenario was run three times, with variations in participants’ initial positions, paths, and speeds to capture real-world complexity.

\begin{figure*}[t]
\centering
\includegraphics[width=1\linewidth]{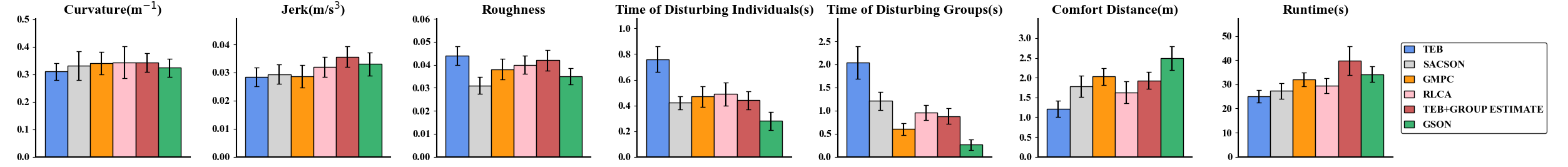}
\caption{\textbf{Real-world experiment results.} \name{} performs comparably on traditional navigation metrics—exhibiting smooth curvature, low roughness, and slightly higher jerk—while excelling in key social navigation metrics. These results highlight its effectiveness in navigating complex social interactions.}

\label{figure:bar_real}
\vspace{-0.4cm}
\end{figure*}

\subsubsection{Baseline and Evaluation Metrics}
We compared \name{}'s performance against several baselines, including classical planners like TEB and TEB enhanced with our group estimation, as well as recent open-source approaches: SACSON~\cite{hirose2023sacson}, an imitation learning-based method; GMPC~\cite{wangGroupbasedMotionPrediction2022}, a group-based motion prediction controller; and RLCA~\cite{long2018towards}, a distributed deep reinforcement learning method. Note that we excluded several LMM-based methods~\cite{payandeh2024social, kong2025autospatial, narasimhan2024olivia, song2024vlm} from the comparison because they are not open source and did not perform reliably in our pilot experiments with our own implementations.
For evaluation, traditional navigation metrics include roughness, curvature, jerk, and runtime~\cite{mavrogiannis2023core,kastnerArenaRosnavDevelopmentBenchmarking2023}. To assess social navigation, we introduced metrics to capture the robot’s impact on individuals and groups. Following prior guidelines~\cite{francisPrinciplesGuidelinesEvaluating2023}, an individual was considered disturbed if the robot approached within 1.2 meters directly in front. For groups, a disturbance was defined by the robot entering the convex hull formed by group members. We measured both the duration of group disturbance and the ``comfort distance''—the average minimum distance between the robot and social groups.

\subsubsection{Result and Analysis}

The results are shown in Fig.~\ref{figure:bar_real}. \name{} performs comparably to other methods in traditional navigation metrics like curvature, jerk, and roughness, achieving smooth and efficient motion with only a slight increase in jerk and \textcolor{black}{runtime. Our speed is slightly lower than the comparison methods, mainly due to two reasons. First, calling the LMM adds at least 1 second of response time. Second, our method actively avoids groups, even if it means taking longer detours. However, the time is not significantly longer since the mid-level planning chooses the optimal path that does not interrupt the group.}
In terms of social navigation, \name{} clearly outperforms baselines. It records the lowest Time of Disturbing Individuals and Groups, showing minimal disruption to pedestrians and social groups. Notably, even compared to TEB+GROUP ESTIMATE, \name{} maintains better group awareness. This is due to the short planning horizon of the TEB planner—when group members are spaced far apart, as in our scenarios, \textit{TEB+GROUP ESTIMATE} can become stuck in local minima.
\name{} also achieves the highest Comfort Distance, maintaining more personal space from social groups—crucial in semantically rich environments. This demonstrates its strong respect for social norms during navigation.
Overall, \name{} balances traditional performance with superior social compliance, making it highly effective for real-world deployment in human-centric environments.

\begin{figure}[t]
\centering
\vspace{-0.2cm}
\includegraphics[width=0.9\linewidth]{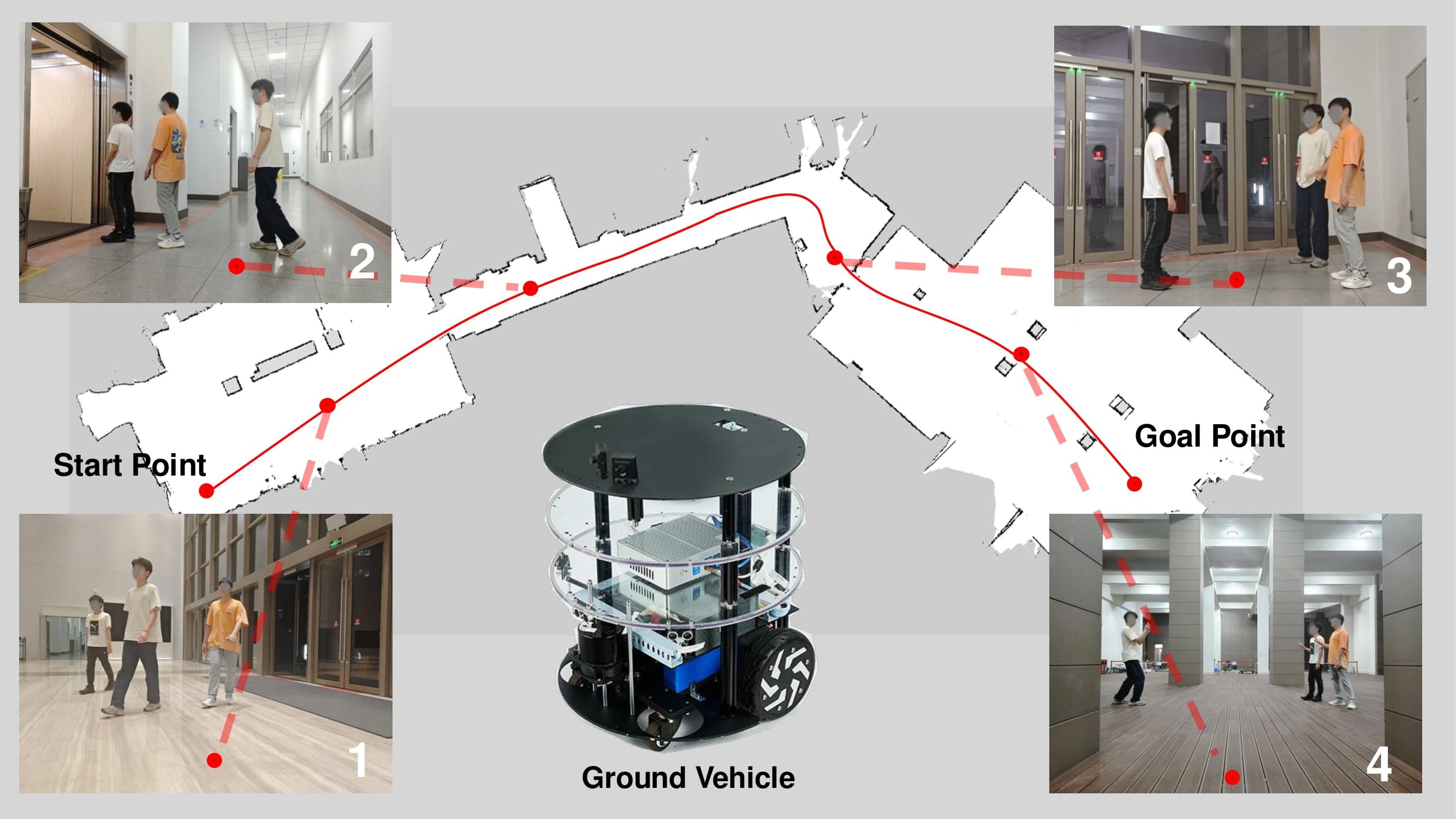}
\caption{The setting of the long-range social navigation task.}
\label{figure:setting}
\vspace{-0.4cm}
\end{figure}

 \begin{figure*}[!t]
\centering
\includegraphics[width=0.9\linewidth]{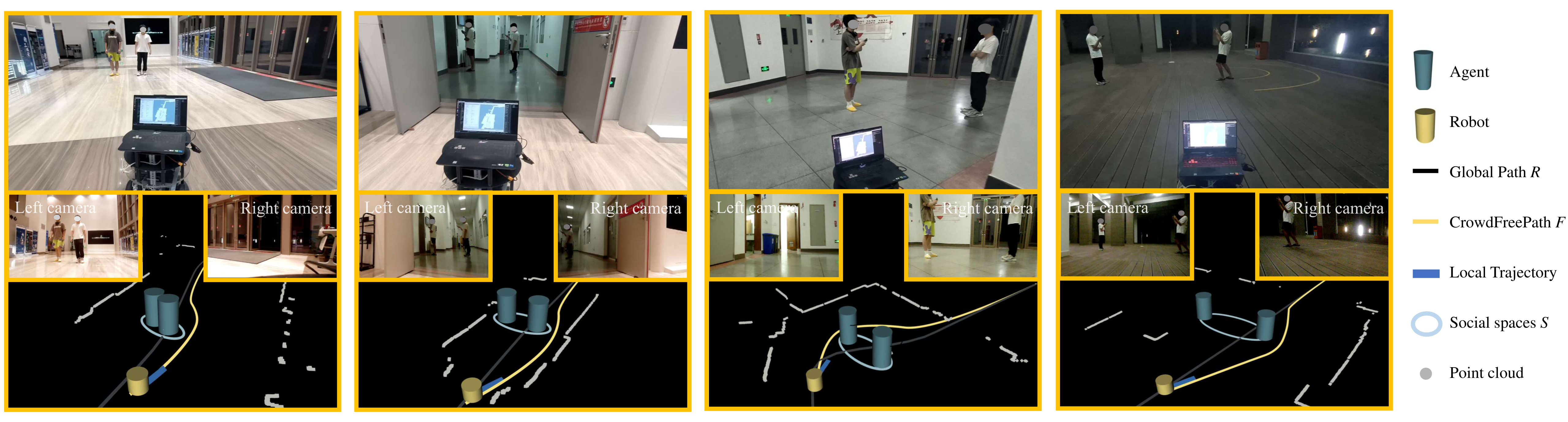}
\caption{\textbf{Demonstration of Real-World Long-Range Experiment.} This figure illustrates the main segment of the long-range real-world experiment, which integrates four smaller scenarios over a nearly 100-meter route from start to finish. The path includes a narrow corridor and an outdoor area with uneven terrain. The robot successfully reached the endpoint while consistently avoiding disturbance to social groups.}
\label{figure:snapshot}
\vspace{-0.5cm}
\end{figure*}

\subsection{Long-Range Social Navigation Demonstration}

We conducted a long-range demonstration in a structured, large-scale building comprising a hall, corridors, and an open square. This extended test integrated four smaller scenarios into a complex setting, where the robot was tasked with identifying social interactions and navigating through these environments (see~\figref{figure:setting}). The start and goal points were defined on opposite sides of the map.

Fig.~\ref{figure:snapshot} shows keyframes from the navigation process. At approximately 10 seconds, the robot encounters a small group of individuals walking ahead. Leveraging its perception and planning modules, it identifies the group and adjusts its trajectory, successfully avoiding the two people.
At around 60 seconds, the robot approaches the entrance of an elevator where a queue has formed. It recognizes the queue and reroutes to avoid cutting through it, ensuring it does not disturb the individuals waiting.
By 146 seconds, while inside a narrow corridor, the robot detects a group engaged in conversation. Based on the group estimation, it computes a seclusive path to bypass the group and maintain smooth navigation through the confined space.
Finally, at around 191 seconds, upon entering an outdoor square, the robot encounters individuals engaged in photography. It identifies their positions and reroutes its path to go around them, ensuring progress while respecting their activity.
In summary, this long-range demonstration validated the robot’s ability to navigate diverse, real-world social environments while maintaining seamless and respectful interactions. Its consistent performance across various challenging scenarios—from avoiding moving individuals to navigating around groups in tight or open areas—highlights its advanced reasoning about social interactions.

\subsection{Group-Aware Social Navigation in Dense Crowds: Limitations and Future Directions}
\textcolor{black}{
\textcolor{black}{
To further investigate the ability of our system in high-density social environments and provide insights, we conducted additional experiments in three real-world scenarios: a university canteen during peak lunch hours, the entrance of a teaching building near an elevator, and the main campus gate. Each involved an average of over 13 pedestrians exhibiting dynamic, diverse social behaviors within the robot's immediate workspace.  
 While the robot successfully completed all navigation tasks, several limitations emerged under dense crowd conditions. We analyze these challenges and offer insights to guide future improvements.}
}

\textcolor{black}{
\noindent\textbf{LMM inference latency}  
Latency from LMM-based group estimation—typically 1.5 to 2 seconds using Qwen2.5-VL-32B—sometimes reduced planner responsiveness to rapidly changing social contexts. To balance reasoning capability and speed, we plan to distill this ability into a smaller model, Qwen2.5-VL-3B, via fine-tuning.
}

\textcolor{black}{
\noindent\textbf{Occlusions and partial observability}  
Egocentric perception suffered from partial occlusions, where individuals blocked one another from the camera’s view, leading to incomplete or inaccurate group detection. In future work, we aim to incorporate predicted bird’s-eye views to infer social relationships beyond egocentric observations.
}

\textcolor{black}{
\noindent\textbf{Complex and adaptive group behavior}
Currently, our framework outputs binary group labels (0 or 1), and the planner conservatively avoids all identified social groups. While sufficient for constructing conservative social spaces, this approach lacks flexibility to adapt to nuanced social contexts (e.g., traversing a queue may be more acceptable than interrupting a group chat). Incorporating finer-grained social understanding is a key next step. One promising direction is to integrate LMM-generated costs—based on different social interaction types—into the planning process.
}

%% file: sections/conclusion.tex
\section{Conclusion}
In this work, we present a novel approach that integrates the visual reasoning capabilities of LMMs with a social structure-aware perception and planning system for mobile robots operating in human-centered environments. 
Our method utilizes the common sense reasoning ability of LMM by applying the visual prompting to predict the social relationships among pedestrians, which allows our planning system to generate socially aware behavior.
Through extensive real-world navigation experiments, our approach demonstrated superior performance and robustness compared to baseline methods, illustrating its potential for improving social awareness and interaction in autonomous robot navigation. Future work could explore extending this framework to more dense social interactions, combining compact state
representation \cite{chen2023llm}, and also distilling knowledge into smaller models for faster inference.